\documentclass[runningheads,a4paper]{llncs}
\usepackage{amsmath}
\usepackage{amssymb}
\usepackage{mathrsfs}
\usepackage{mathtools}
\usepackage{booktabs}
\usepackage{color}
\usepackage{array}
\usepackage{graphicx}
\usepackage[utf8]{inputenc}

\usepackage{mynotation}

\newcommand{\keywords}[1]{\par\addvspace\baselineskip
	\noindent\keywordname\enspace\ignorespaces#1}

\title{DCCO: Towards Deformable Continuous Convolution Operators for Visual Tracking}

\author{Joakim Johnander, Martin Danelljan, Fahad Shahbaz Khan, Michael Felsberg}
\authorrunning{J.~Johnander, M.~Danelljan, F.S.~Khan, M.~Felsberg}

\institute{Computer Vision Laboratory, Dept.\ of Electrical Engineering, Link\"{o}ping University}

\begin{document}


\mainmatter  

\maketitle

\begin{abstract}
Discriminative Correlation Filter (DCF) based methods have shown competitive performance on tracking benchmarks in recent years. Generally, DCF based trackers learn a rigid appearance model of the target. However, this reliance on a single rigid appearance model is insufficient in situations where the target undergoes non-rigid transformations. In this paper, we propose a unified formulation for learning a deformable convolution filter. In our framework, the deformable filter is represented as a linear combination of sub-filters. Both the sub-filter coefficients and their relative locations are inferred jointly in our formulation. Experiments are performed on three challenging tracking benchmarks: OTB-2015, TempleColor and VOT2016. Our approach improves the baseline method, leading to performance comparable to state-of-the-art.

\end{abstract}

\section{Introduction}
Generic visual object tracking is the computer vision problem of estimating the trajectory of a target throughout an image sequence, given only the initial target location. Visual tracking is useful in numerous applications, including autonomous driving, smart surveillance systems and intelligent robotics. The problem is challenging due to large variations in appearance of the target and background, as well as challenging situations involving motion blur, target deformation, in- and out-of-plane rotations, and fast motion.

To tackle the problem of visual tracking, several paradigms exist in literature \cite{VOT2016}. Among different paradigms, approaches based on the Discriminative Correlation Filters (DCF) based framework have achieved superior results, evident from recent the Visual Object Tracking (VOT) challenge results \cite{VOT2015}\cite{VOT2016}. This improvement in performance, both in terms of precision and robustness, is largely attributed to the use of powerful multi-dimensional features such as HOG, Colornames, and deep features \cite{DanelljanCVPR2016a}\cite{HCF_ICCV15}\cite{DanelljanCVPR14}, as well as sophisticated learning models \cite{DanelljanICCV2015}\cite{DanelljanECCV2016}.

Despite the improvement in tracking performance, the aforementioned state-of-the-art DCF based approaches employ a single rigid model of the target. However, this reliance on a single rigid model is insufficient in situations involving rotations and deformable targets. In such complex situations, the rigid filters fail to capture information of the target parts that move relative to eachother. This desired information can be retained by integrating deformability in the DCF filters. Several recent works aim at introducing part-based information into the DCF framework \cite{liu2015real}\cite{li2015reliable}\cite{lukevzivc2016deformable}. These approaches introduce an explicit component to integrate the part-based information in the learning. Different to these approaches, we investigate a deformable DCF  model, which can be learned in unified fashion.

In many real-world situations, such as a running human or a rotating box, different regions of the target deform relative to each other. Ideally, such information should be integrated in the learning formulation by allowing the regions of the appearance model to deform accordingly. This flexibility in the tracking model reduces the need of highly invariant features, thereby increasing the discriminative power of the model. However, increasing the flexibility  and complexity of the model introduces the risk of over-fitting and complex inference mechanisms, which degrades the robustness of the tracker. In this paper, we therefore advocate a unified formulation, where the deformable filter is learned by optimizing a single joint objective function. Additionally, this unified strategy enables the careful incorporation of regularization models to tackle the risk of over-fitting.

\begin{figure}[!t]
  \centering
  \includegraphics[width=0.25\textwidth]{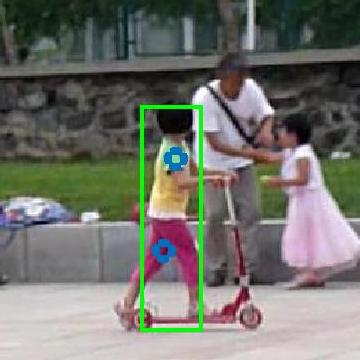}%
  \includegraphics[width=0.25\textwidth]{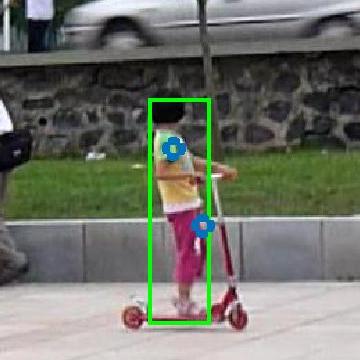}%
  \includegraphics[width=0.25\textwidth]{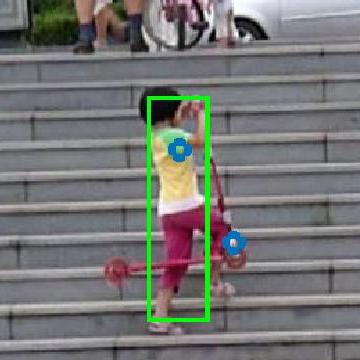}%
  \includegraphics[width=0.25\textwidth]{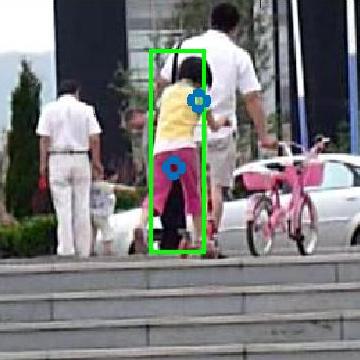}
  \includegraphics[width=0.25\textwidth,trim={7cm 5cm 7cm 6cm},clip]{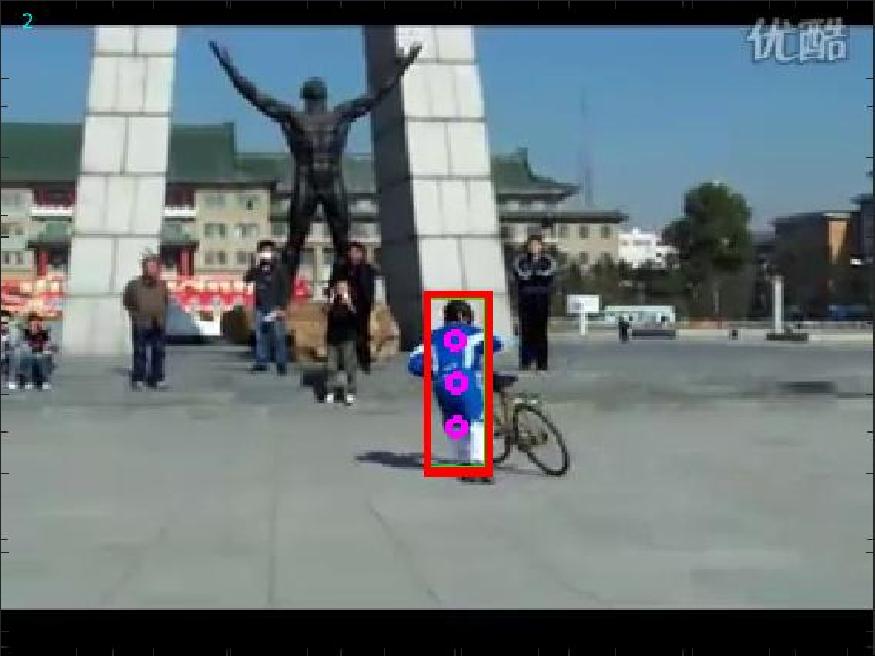}%
  \includegraphics[width=0.25\textwidth,trim={7cm 5cm 7cm 6cm},clip]{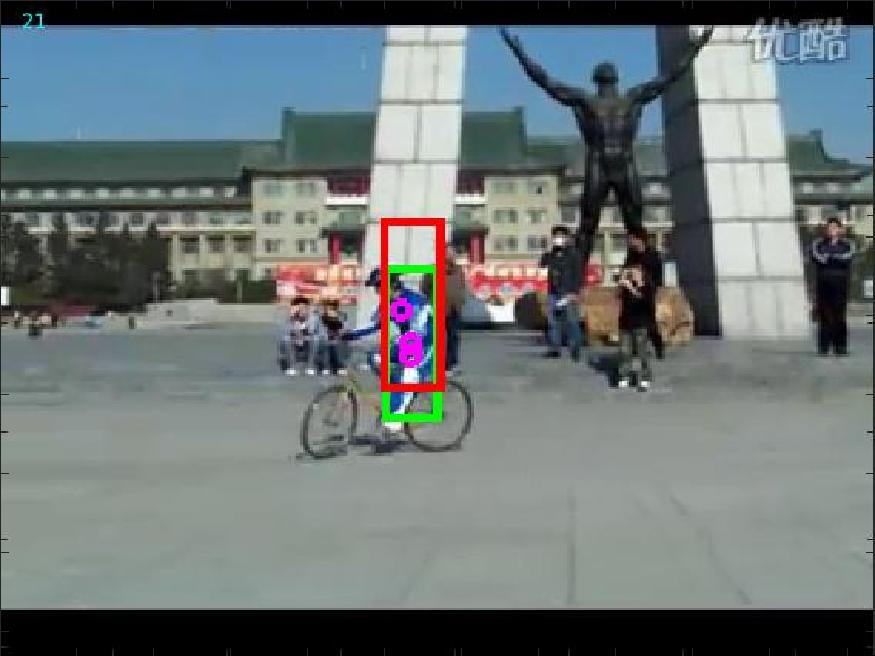}%
  \includegraphics[width=0.25\textwidth,trim={7cm 5cm 7cm 6cm},clip]{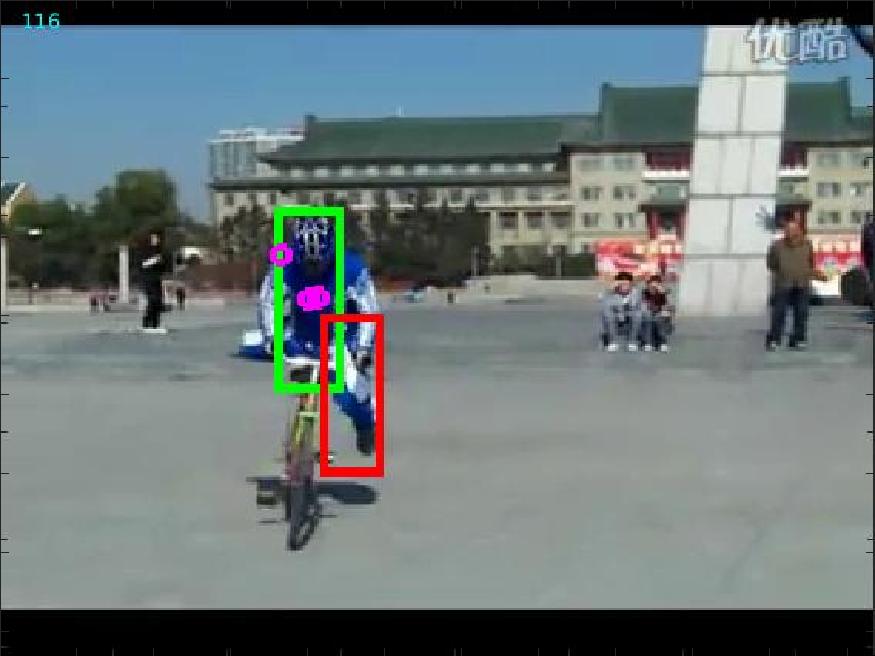}%
  \includegraphics[width=0.25\textwidth,trim={7cm 5cm 7cm 6cm},clip]{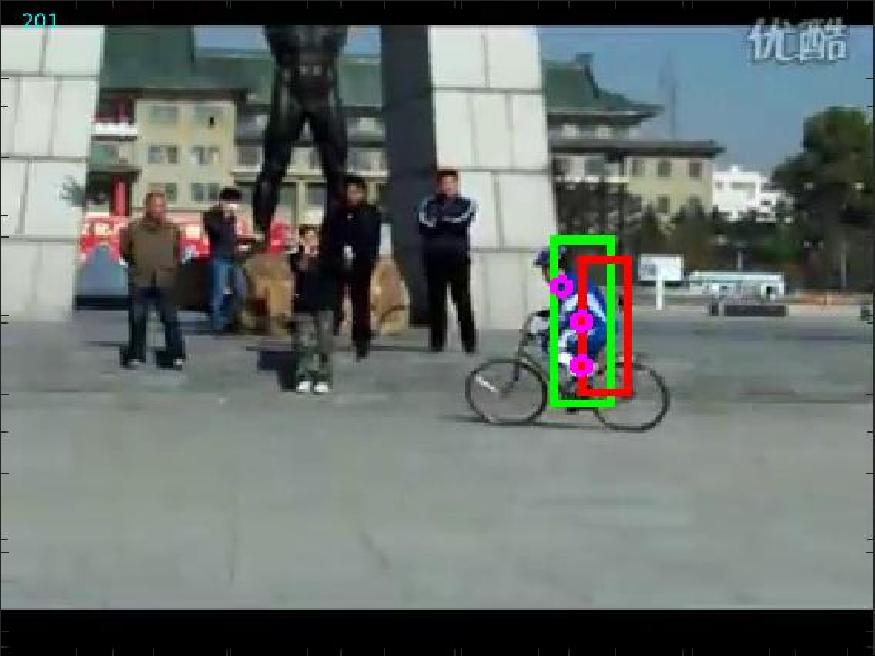}
  \includegraphics[width=0.25\textwidth,trim={10cm 8cm 10cm 8cm},clip]{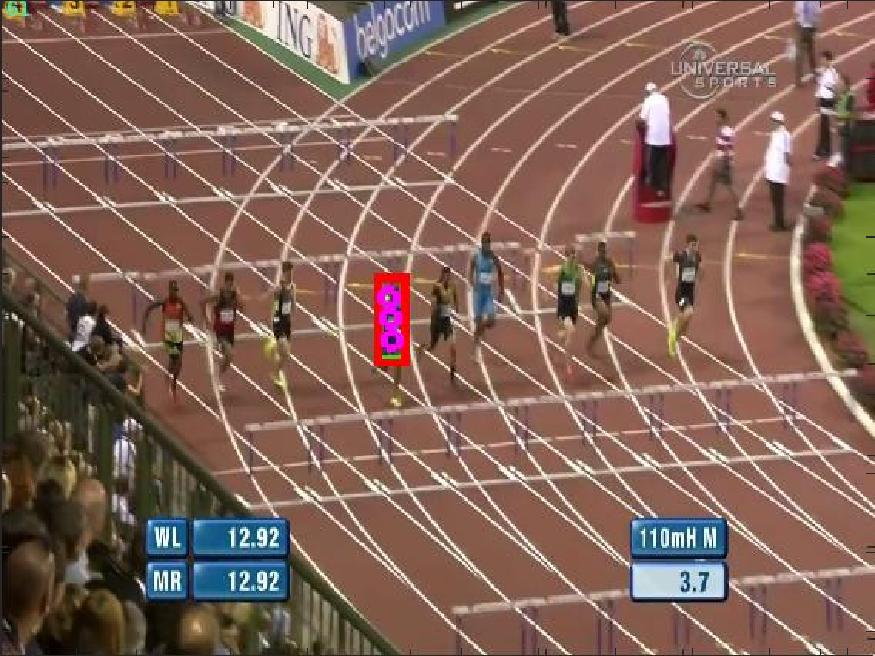}%
  \includegraphics[width=0.25\textwidth,trim={10cm 8cm 10cm 8cm},clip]{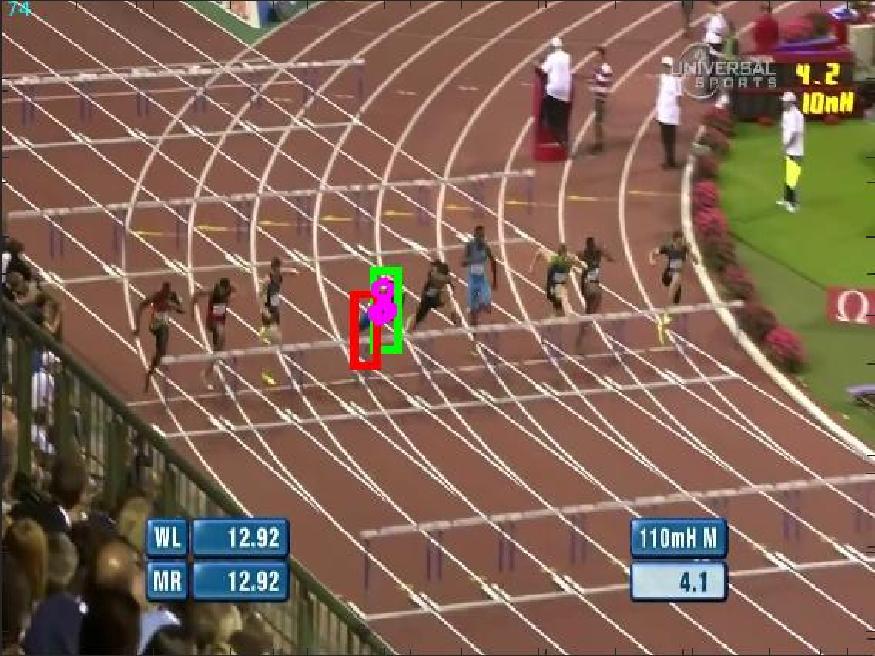}%
  \includegraphics[width=0.25\textwidth,trim={10cm 8cm 10cm 8cm},clip]{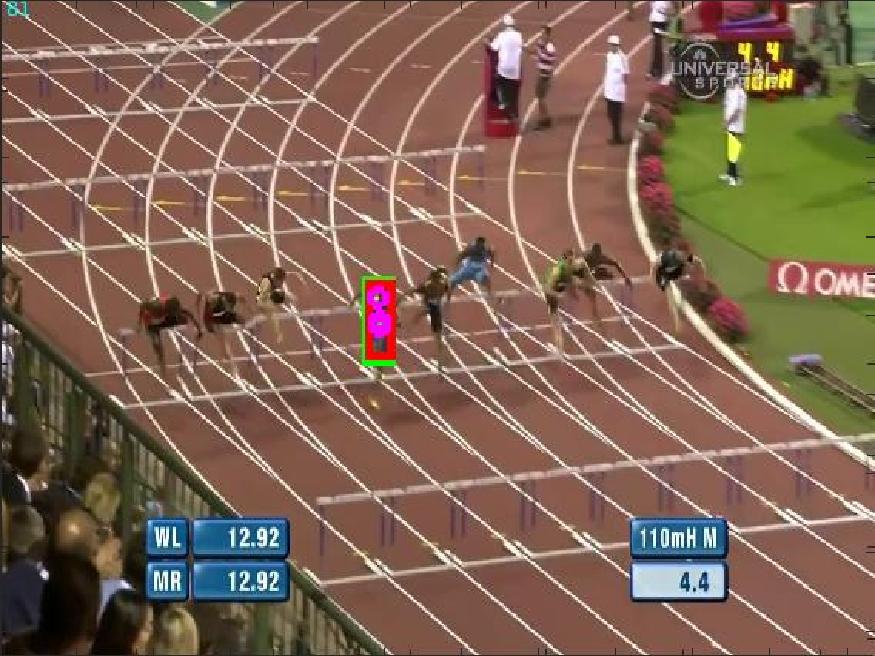}%
  \includegraphics[width=0.25\textwidth,trim={10cm 8cm 10cm 8cm},clip]{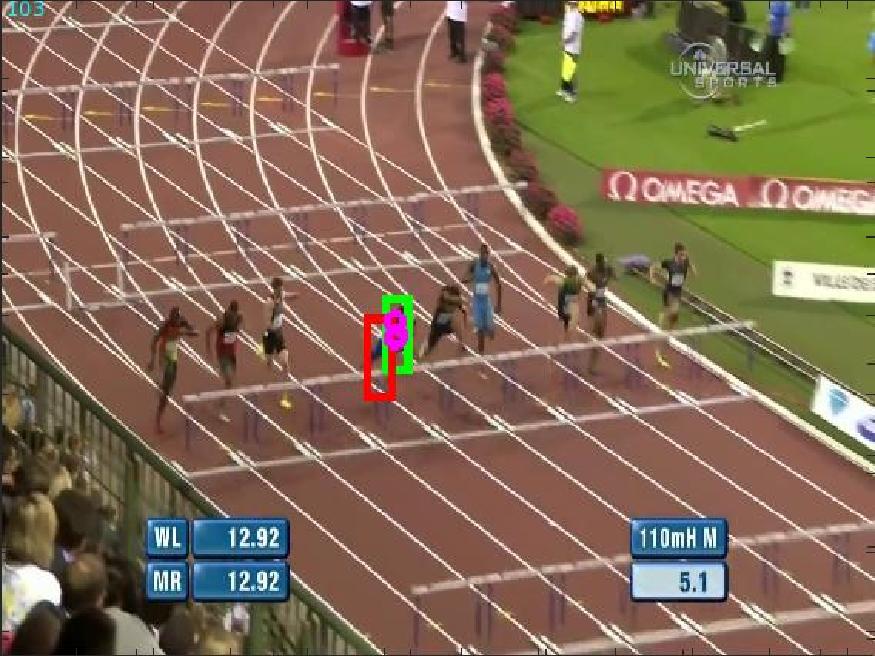}
  \caption{
  	Example tracking results of our deformable correlation filter approach on three challenging sequences. The circles mark sub-filter locations and the green box is the predicted target location. The red boxes (in the middle and lower rows) show the baseline predictions. The sub-filter locations deform according to the appearance changes of the target in the presence of deformations. 
  }
  \vspace{-0.5cm}
  \label{fig:examples}
\end{figure}

\subsubsection{Contribution}
We propose a unified framework for learning a deformable convolution filter in a discriminative fashion. The deformable filter is represented as a linear combination of sub-filters. The deformable filter is learned by jointly optimizing the sub-filter coefficients \emph{and} their relative locations. To avoid over-fitting, we propose to regularize the sub-filter locations with an affine deformation model. We further derive an efficient online optimization procedure to infer the parameters of the model. Experiments on three challenging tracking benchmarks suggest that our method improves the performance in challenging situations.

\section{Related Work}
In recent years, Discriminative Correlation Filters (DCF) based tracking methods have shown competitive performance in terms of accuracy and robustness on tracking  benchmarks \cite{VOT2016}\cite{OTB2015}. In particular, the success of DCF based methods is evident from the outcome of the Visual Object Tracking (VOT) 2014 and 2016 challenges \cite{VOT2016} where the top-rank trackers employ variants of the DCF framework. In DCF framework, a correlation filter is learned from a set of training samples to discriminate between the target and background appearance. The training of the filter is performed in a sliding-window manner by exploiting the properties of circular correlation. The original DCF based tracking approach by Bolme et al. \cite{MOSSE2010} was restricted to a single feature channel and was later extended to multi-channel feature maps \cite{galoogahiICCV13}\cite{DanelljanCVPR14}\cite{Henriques14}. Most recent  advancement in DCF based tracking  performance is attributed to including scale estimation \cite{DanelljanBMVC14}\cite{Li2014}, deep features \cite{DanelljanVOT2015}\cite{HCF_ICCV15}, spatial regularization \cite{DanelljanICCV2015}, and continuous convolution filters \cite{DanelljanECCV2016}.

Several recent works have shown that integrating the part-based information improve the tracking performance. The work of \cite{liu2015real} introduces a part-based approach where each part utilizes the kernalized correlation filter (KCF) tracker and argues that partial occlusions can effectively be handled by adaptive weighting of the parts. The work of \cite{li2015reliable} tracks several patches, each with a KCF, by fusing the information using a particle filter to estimate position, width and height. Lukezic et. al. \cite{lukevzivc2016deformable} introduces a sophisticated model with several parts held together by a spring-like system by minimizing an energy function based on the part-filter responses.

\noindent\textbf{Our approach:} Different to aforementioned approaches, we propose a theoretical framework by designing a single deformable correlation filter. In our approach, the coefficients and locations of all sub-filters are learned jointly in a \emph{unified} framework. Additionally, we integrate our deformable correlation filter in a recently introduced state-of-the-art DCF tracking framework \cite{DanelljanECCV2016}.

\section{Continuous Convolution Operators for Tracking}
\label{sec:CCOT}
In this work, we propose a deformable correlation tracking formulation. As a starting point, we use the recent Continuous Convolution Operator Tracker (C-COT) formulation \cite{DanelljanECCV2016} due to two main advantages compared to current template based correlation filter trackers. Firstly, the continuous reformulation of the learning problem benefits from a natural integration of multi-resolution deep features and continuous-domain score map predictions. Secondly, it provides an efficient optimization framework based on the Conjugate Gradient method. For efficiency, we also employ components of its descendant tracker \emph{ECO} \cite{DanelljanCVPR2017}.

For a given target object in a video, the C-COT discriminatively learns a convolution filter $f$ that acts as an instance-specific object detector. Different from previous approaches, the filter $f$ is viewed as a continuous function represented by its Fourier series coefficients. The detection scores are computed by first extracting a $D$-dimensional feature map $x$ from the local image region of interest. Typically, the sample $x$ consists of HOG or multi-resolution deep convolutional features. We let $x_d[n_1,n_2]$ denote the value of the $d$-th feature channel at the spatial location $(n_1,n_2)$ in the feature map. The continuous scores in the corresponding image region are determined by the convolution operation $S_f\{x\} = \sum_{d=1}^D f_d \conv J^d\{x_d\}$, where $J^d\{x_d\}$ is an interpolation operator mapping the samples from the discrete to the continuous domain.

The filter $f$ is trained in a supervised fashion, given a set of sample feature maps $\{x^1,x^2,\dots,x^C\}$ and corresponding label score maps $\{y^1,y^2,\dots,y^C\}$, by minimizing the objective,
\begin{equation}
	\label{eq:ccot_training}
  \epsilon(f) = \sum_{c=1}^C\alpha^c\|S_f\{x^c\} - y^c\|^2 + \sum_{d=1}^D\|w^d\cdot f_d\|^2 .
\end{equation}
The first term penalizes classification errors of each sample using the squared $L^2$-norm. The sample $c$ is weighted by the positive weight factor $\alpha^c$, which is typically set using a learning rate parameter. The second term deploys a continuous spatial regularization function $w^d$, that penalizes high magnitude filter coefficients to alleviate the periodic boundary effects. Element-wise multiplication is denoted as $\cdot$. The label score function $y^c$ is generally set to a Gaussian function with a narrow peak at the target center. Note that a sample feature map $x^c$ contains both target appearance and the surrounding background. The filter is hence trained to predict high activation scores at the target center and low scores at the neighboring background. In practice, training and detection is performed directly in the Fourier domain, utilizing the FFT algorithm and the convolution properties of the Fourier series.

As related methods, the C-COT method works in two main steps. (i) When a new sample is received, the target position and scale are estimated, i.e. $S_f\{x\}$ is calculated using the estimated filter $f$ for different scales using a scale pyramid. The new target state is then estimated as the position and scale that maximizes the detection score. (ii) To update the model, a sample $(x^c,y^c)$ is first added to the training set, where $x^c$ is extracted in the estimated target scale. The filter is then refined by minimizing the objective \eqref{eq:ccot_training}. This is done by using conjugate gradient to solve the arising normal equations. We refer to \cite{DanelljanECCV2016} for further details. To enhance the efficiency of the tracker, we further deploy the factorized convolution approach and update strategy recently proposed in \cite{DanelljanCVPR2017}.

\section{Method}

Here, we introduce a deformable correlation filter tracking model. A classic DCF contains an assumption that the target is rigid and will not rotate. The filter can handle violations to this assumption if a significant part of the target still fulfills it, or by using features with sufficient invariance. Examples of such model violations are sequences showing humans running or a change of perspective. By dividing the filter into sub-filters which can move relative to each other, they can fit more accurately onto a smaller part of the target. A standard DCF may choose to discard or weigh down information about a moving part whereas our approach allows one sub-filter to focus on this information explicitly, and move with that part. By writing the filter as a linear combination of sub-filters we can optimize a joint loss over all the sub-filter coefficients and the sub-filter positions jointly.

\subsection{Deformable Correlation Filter}
We construct a deformable convolution filter as a linear combination of trainable sub-filters. The filter becomes deformable by allowing the relative locations of the filters to change along to the target transformations. Formally, we denote the sub-filter with $f^m$ and let $p^{c,m}=(p_1^{c,m},p_2^{c,m})$ be its relative location in the frame $c$. The filter $f$ at frame $c$ is obtained as a linear combination of the shifted sub-filters,
\begin{equation}
	\label{eq:def_filter}
  f(t_1,t_2) = \sum_{m=1}^M f^m(t_1-p_1^{c,m},t_2-p_2^{c,m}).
\end{equation}
We jointly learn both the sub-filter coefficients $f^m$ and their locations $p^{c,m}$ by minimizing a joint loss.,
\begin{equation}
  \label{lossdeform}
  \epsilon(f,p) = \epsilon_1(f,p) + \epsilon_2(f) + \epsilon_3(p),
\end{equation}
where each term is described below.

\subsubsection{Classification Error}
The loss for the discrepancy between the desired response and the filter response for sample $x^c$ is
\begin{equation}
  \epsilon_1(f,p) = \sum_{c=1}^C\alpha^c\|S_f\{x^c\} - y^c\|^2 ,
\end{equation}
where $\alpha^c$ is the weight for sample $c$. From the translation invariance of the convolution operation and the definition \eqref{eq:def_filter}, the classification scores can be computed as, 
\begin{equation}
S_f\{x^{c}\}(t_1,t_2) = \sum_{m=1}^M S_{f^m}\{x^{c}\}(t_1-p_1^{c,m},t_2-p_2^{c,m}).
\end{equation}
The score operator $S_{f^m}\{x^c\}$ is defined as described in section~\ref{sec:CCOT}.

\subsubsection{Spatial Regularization}
A spatial regularization of the filters enforces low filter coefficients close to the edges,
\begin{equation}
  \epsilon_2(f) = \sum_{m=1}^M\sum_{d=1}^D\|w^{m,d}\cdot f_d^m\|^2,
\end{equation}
where $w^{m,d}$ is the continuous spatial regularization function for filter $m$. We assume different spatial regularization functions for the different sub-filters as it may be desireable for the sub-filters to track regions of different size. In our experiments, by using two different spatial regularizations where one is much tighter, we let one sub-filter track the whole target while the others track smaller patches. Please note that $\epsilon_2(f)$ does not depend on the sub-filter positions.

\subsubsection{Regularization of Sub-filter Positions}
To regularize the sub-filter positions, we add a deformable model that incorporates prior information of typical target deformations. In this work, we use a simple yet effective model, namely that the current sub-filter positions are related to their initial positions by a linear mapping. The resulting regularization term is thus given by,
\begin{equation}
  \epsilon_3(p) = \lambda_p\sum_{m=1}^M\|p^{c,m} - Rp^{1,m}\|^2 .
\end{equation}
Here, $p^{c,m}$ is the position of sub-filter $m$ in frame $c$, and $R\in\reals^{2\times2}$ is a transformation matrix. In our experiments we use a full linear transform, which is optimized jointly during the learning. $\lambda_p$ is a parameter determining the regularization impact. This part of the loss does not depend on the sub-filter coefficients.

\subsection{Fourier Domain Formulation}
The optimization is performed in the Fourier domain using Parseval's formula. This results in a finite representation of the continuous filters using truncated Fourier series.

Let $\hat{\cdot}$ denote the Fourier coefficients for any given, sufficiently nice function. By linearity of the Fourier transform
\begin{equation}
  \widehat{S_f\{x^c\}}[k_1,k_2] = \sum_{m=1}^M\beta[k_1,k_2]\widehat{S_{f^m}\{x^c\}}[k_1,k_2]
\end{equation}
where
\begin{equation}
  \beta[k_1,k_2] = e^{-i2\pi p_1^{c,m}k_1/T_1}e^{-i2\pi p_2^{c,m}k_2/T_2}
\end{equation}
and
\begin{equation}
\widehat{S_{f^m}\{x^c\}}[k_1,k_2] = \left(\sum_{d=1}^D\hat{f}_d^m[k_1,k_2]\widehat{J^d\{x^c\}}[k_1,k_2]\right).
\end{equation}

Given $C$ samples, we optimize the filter in the C-COT framework. The objective \ref{lossdeform} is minimized by using Parseval's formula. We get the corresponding objective
\begin{equation}
  \epsilon(f,p) = \sum_{c=1}^C\alpha^c\|\widehat{S_{f}\{x^c\}} - \hat{y}^c|^2 + \sum_{m=1}^M\sum_{d=1}^D\|\hat{w}^{m,d}*\hat{f}_d^m\|^2 + \lambda_p\sum_{m=1}^M\|p^{c,m} - Rp^{1,m}\|^2
\end{equation}
which will be minimized by an alternate optimization strategy where we iteratively update the sub-filter coefficients and positions.

\subsection{Updating the Filter Coefficients}
The Fourier coefficients are truncated such that for feature dimension $d$ only the $K^d$ first coefficients are used (resulting in $2K^d+1$ coefficients in total for that dimension). Also define $K = \max_d K^d$. To minimize the functional we rewrite it as a least squares problem which can be solved via its normal equations. The normal equations are then solved using conjugate gradient. Let $\cdot^H$ be the conjugate transpose. We define a block matrix with $C\times MD$ blocks
\begin{equation}
  A = \begin{pmatrix}
    A^1\\
    \vdots\\
    A^C
    \end{pmatrix},\quad
  A^c = \begin{pmatrix}A^{c,1} & \dots & A^{c,M}\end{pmatrix},\quad
  A^{c,m} = \begin{pmatrix}A^{c,m,1} & \dots & A^{c,m,D}\end{pmatrix}
\end{equation}
where $A^{c,m,d}$ is a diagonal matrix of size $K\cdot K\times K^d\cdot K^d$
\begin{equation}
  A^{c,m,d} = \text{diag}\begin{pmatrix}
    \beta[-K^d,-K^d]\widehat{J^d\{x^c\}}[-K^d,-K^d]\\
    \vdots\\
    \beta[-K^d,K^d]\widehat{J^d\{x^c\}}[-K^d,K^d]\\
    \vdots\\
    \beta[K^d,K^d]\widehat{J^d\{x^c\}}[K^d,K^d]\\
  \end{pmatrix}.
\end{equation}
Further define
\begin{equation}
  \vechat{f} = \begin{pmatrix}
    \vechat{f}^1\\
    \vdots\\
    \vechat{f}^M
  \end{pmatrix},\quad
  \vechat{f}^m = \begin{pmatrix}
    \vechat{f}_1^{m}\\
    \vdots\\
    \vechat{f}_D^{m}
  \end{pmatrix},\quad
  \vechat{f}_d^{m} = \begin{pmatrix}
    f_d^{m}[-K^d,-K^d]\\
    \vdots\\
    f_d^{m}[-K^d,K^d]\\
    \vdots\\
    f_d^{m}[K^d,K^d]
  \end{pmatrix}
\end{equation}
and
\begin{equation}
  \vechat{y} = \begin{pmatrix}
    \vechat{y}^1\\
    \vdots\\
    \vechat{y}^C
  \end{pmatrix}.
\end{equation}
Lastly, let $\Gamma$ denote a diagonal matrix containing the learning rate $\alpha^c$, of size $CK\times CK$; and $W$ denote a Toeplitz matrix corresponding to summation of the convolutions with $w^{m,d}$. Using these definitions the objective becomes
\begin{equation}
  \epsilon(f,p) = \sum_{c=1}^C\alpha^c\|A^c\vechat{f} - \vechat{y}^c\|^2 + \|W\vechat{f}\|^2 + \epsilon_3(p).
\end{equation}
We discard $\epsilon_3(p)$ while minimizing the objective over $f$, as it will be addressed in the next step. The objective is then minimized by solving
\begin{equation}
  (A^H\Gamma A + W^HW)\vechat{f} = A^H\vechat{y}
\end{equation}
using the method of conjugate gradient.

\subsection{Displacement Estimation of the Sub-Filters}
The sub-filters are moved by minimizing the objective with respect to the sub-filter positions. This problem is not convex, and we resort to gradient descent utilizing Barzilai-Borwein's method \cite{barzilai1988two}. The perk of their method is that the steplength is adaptive. The gradient is found as
\begin{equation}
  \frac{d}{dp^{c,m}}\epsilon(f) = \frac{d}{dp^{c,m}}\epsilon_1(f) + \frac{d}{dp^{c,m}}\epsilon_3(p)
\end{equation}
where
\begin{equation}
  \frac{d}{dp^{c,m}}\epsilon_1(f) = 2(\widehat{S_f\{x^c\}}-\hat{y}^c)e^{-i2\pi p_1^{c,m}k_1/T_1}e^{-i2\pi p_2^{c,m}k_2/T_2}\widehat{S_{f^m}\{x^c\}}\begin{pmatrix}
    -i2\pi k_1/T\\
    -i2\pi k_2/T
  \end{pmatrix}
\end{equation}
and
\begin{equation}
  \frac{d}{dp^{c,m}}\epsilon_3(p) = 2\lambda_{\text{p}}(p^{c,m} - Rp^{1,m}).
\end{equation}
Note that $\epsilon_2(f)$ does not depend on the sub-filter positions, and hence the derivative with respect to the sub-filter positions is zero. In our experiments we let $R$ be either the identity matrix, or an affine transform. The translation part of the affine transform is handled during the target position estimation described in section 3. Hence the affine transform can be considered equivalent to a linear transform. The linear transform is estimated in each step of gradient descent using a closed form expression. This is done by rewriting the problem as an over-determined linear system of equations and solve it via its normal equations.

\section{Experiment and Results}
We validate our approach by performing comprehensive experiments on three tracking benchmarks: OTB-2015 \cite{OTB2015}, TempleColor \cite{TempleColor} and VOT2016 \cite{VOT2016}.

\begin{table}[!t]
  \caption{Baseline comparison on the OTB-2015 dataset with the two different regularizations of the sub-filter positions. The affine transform provides the best results.}
  \label{table:baselinetransf}
  \centering
    \begin{tabular}{|l||c|c|c|}
      \hline
      &Baseline, no deformability&Affine &Identity\\\hline\hline
      Mean OP&83.2&\textbf{\textcolor{red}{83.9}}&\textit{\textcolor{blue}{83.4}}\\\hline
      Mean AUC&68.4&\textbf{\textcolor{red}{69}}&\textit{\textcolor{blue}{68.5}}\\\hline
    \end{tabular}
    \vspace{-0.0cm}
\end{table}

\begin{table}[!t]
  \caption{Baseline comparison on the OTB-2015 dataset when using different set of features for the sub-filters.}
  \label{table:baselinefeat}
  \resizebox{\columnwidth}{!}{
    \begin{tabular}{|l||c|c|c|c|c|c|c|}
      \hline
      &Baseline&Shallow $+$ CN&Shallow&Shallow $+$ Deep&Deep&CN\\\hline\hline
      Mean OP&83.2&83.6&83.5&83.6&\textit{\textcolor{blue}{83.9}}&\textbf{\textcolor{red}{83.9}}\\\hline
      Mean AUC&68.4&\textbf{\textcolor{red}{69}}&\textit{\textcolor{blue}{68.9}}&\textit{\textcolor{blue}{68.9}}&\textbf{\textcolor{red}{69}}&68.8\\\hline
    \end{tabular}
  }
\end{table}

\subsection{Implementation Details}
In our experiments we employ two types of features: Color Names, and ``Deep Features'' extracted from the Convolutional Neural Network (CNN). We use the network VGG-m and extract features from the layers Conv-1 and Conv-5. We use different number of sub-filters depending on the target size. We employ a ``root-filter'' which is a subfilter that is always centered around the target and utilizes both shallow features and deep features from a CNN. The locations of the sub-filters are continuously updated and has a strong regularization to enforce locality. We test different feature sets for these sub-filters. The sub-filters are initialized in the first frame where they are placed in a grid. We use $\lambda_P = 3\cdot 10^{-6}$ on VOT2016 and TempleColor datasets, and use $\lambda_P = 3\cdot10^{-4}$ on the OTB-2015 dataset. We use the same set of parameters for all videos in each dataset.

\subsection{Baseline Comparison}
We perform baseline comparisons on the OTB-2015 dataset with 100 videos. We compare different features for the sub-filters, and different regularization for their positions. We evaluate the tracking performance in terms of mean overlap precision (OP) and area-under-the-curve (AUC). The overlap precision (OP) is calculated as the fraction of frames in the video where the intersection-over-union (IoU) overlap with the ground truth exceeds a threshold of 0.5 (PASCAL criterion). The area-under-the-curve (AUC) is calculated from the success plot where the mean OP is plotted over the range of IoU thresholds over all videos.

\begin{figure}[!t]
  \centering
  \includegraphics[width=0.5\textwidth]{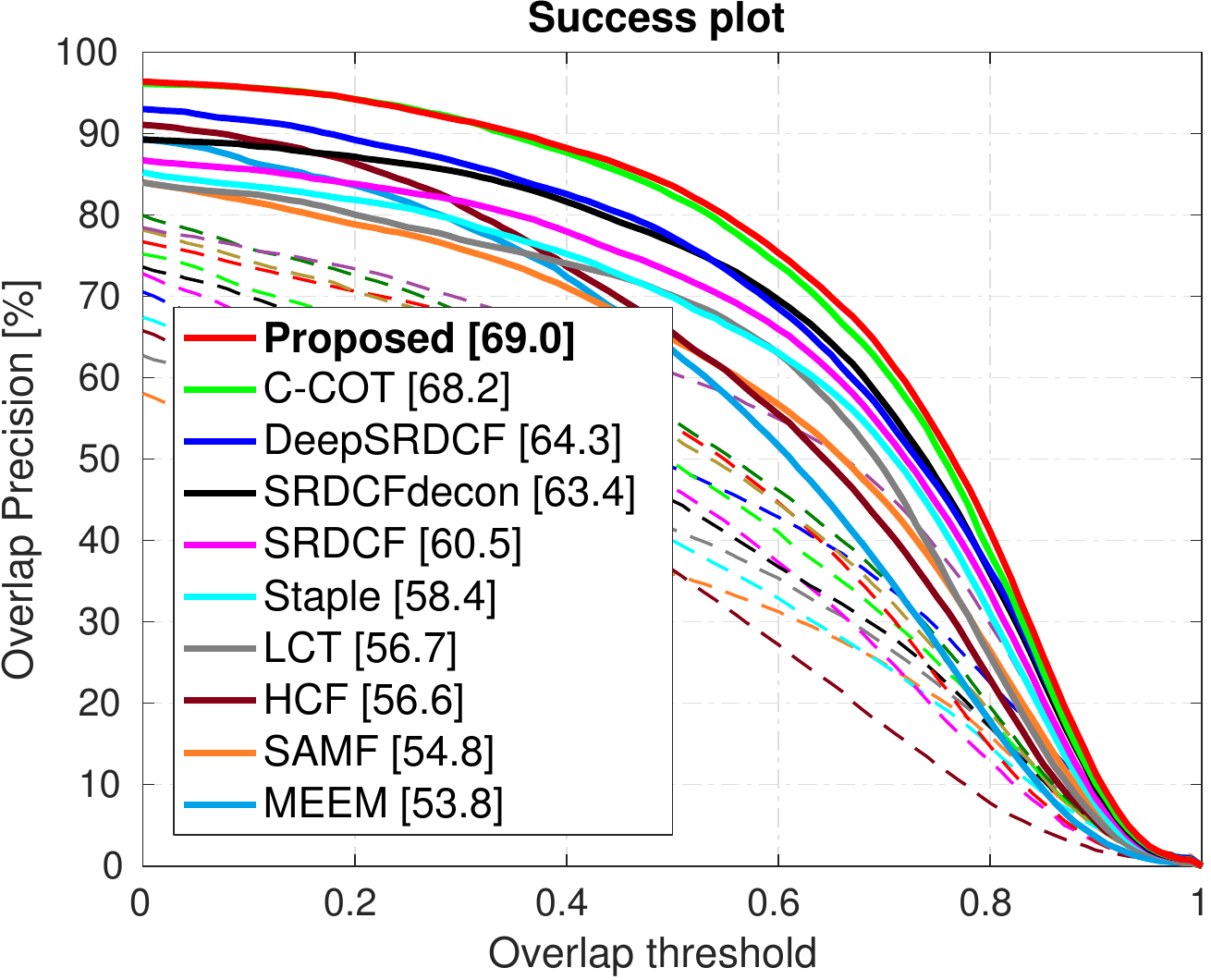}%
  \includegraphics[width=0.5\textwidth]{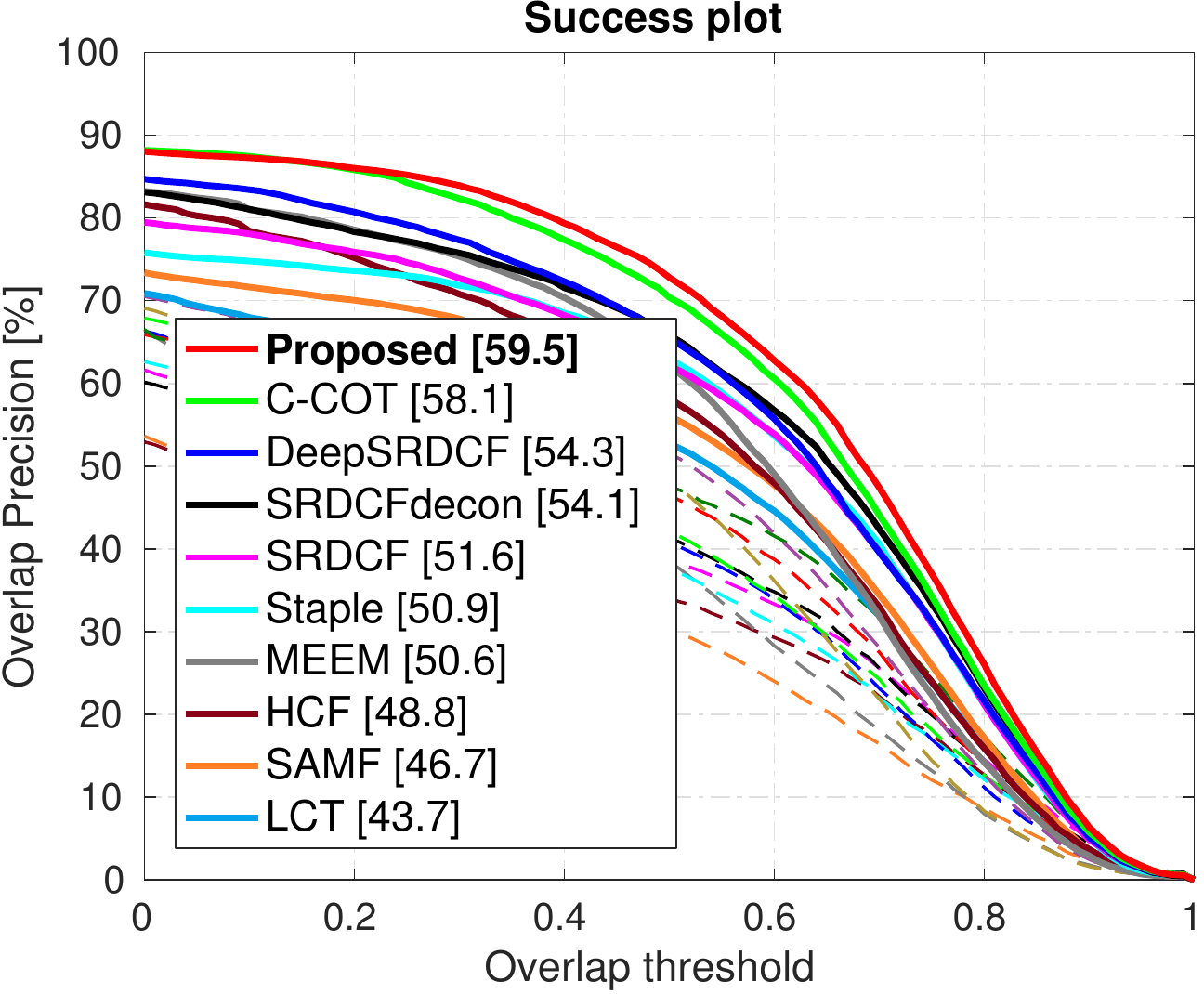}
  \caption{Success plots on the OTB-2015 (left) and TempleColor (right) datasets, compared to state-of-the-art. The AUC score of each tracker is shown in the legend. We show slight performance increases on both datasets.}
  \label{fig:results}
\end{figure}
  
\begin{figure}[!t]
  \centering
  \includegraphics[width=0.33\textwidth]{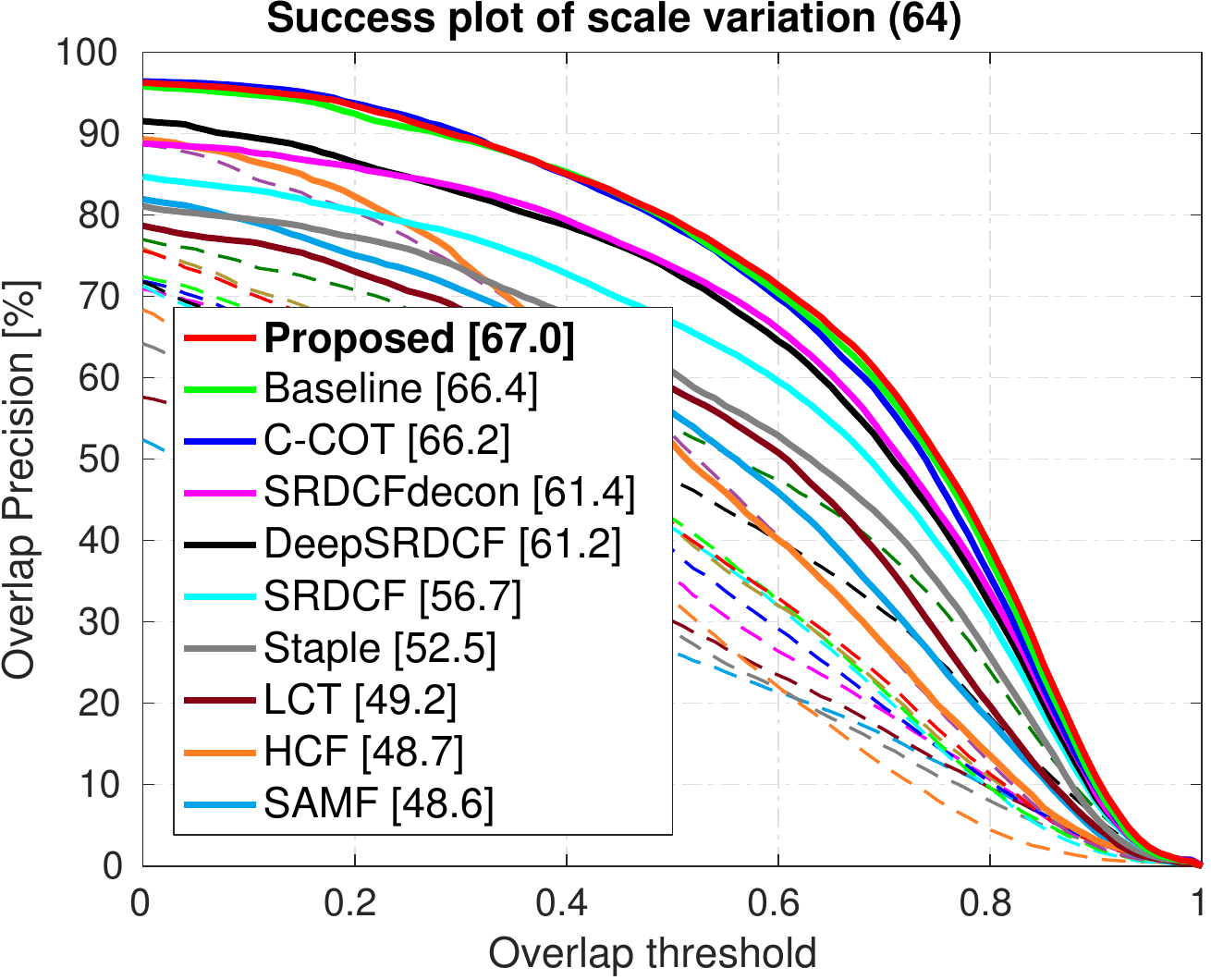}%
  \includegraphics[width=0.33\textwidth]{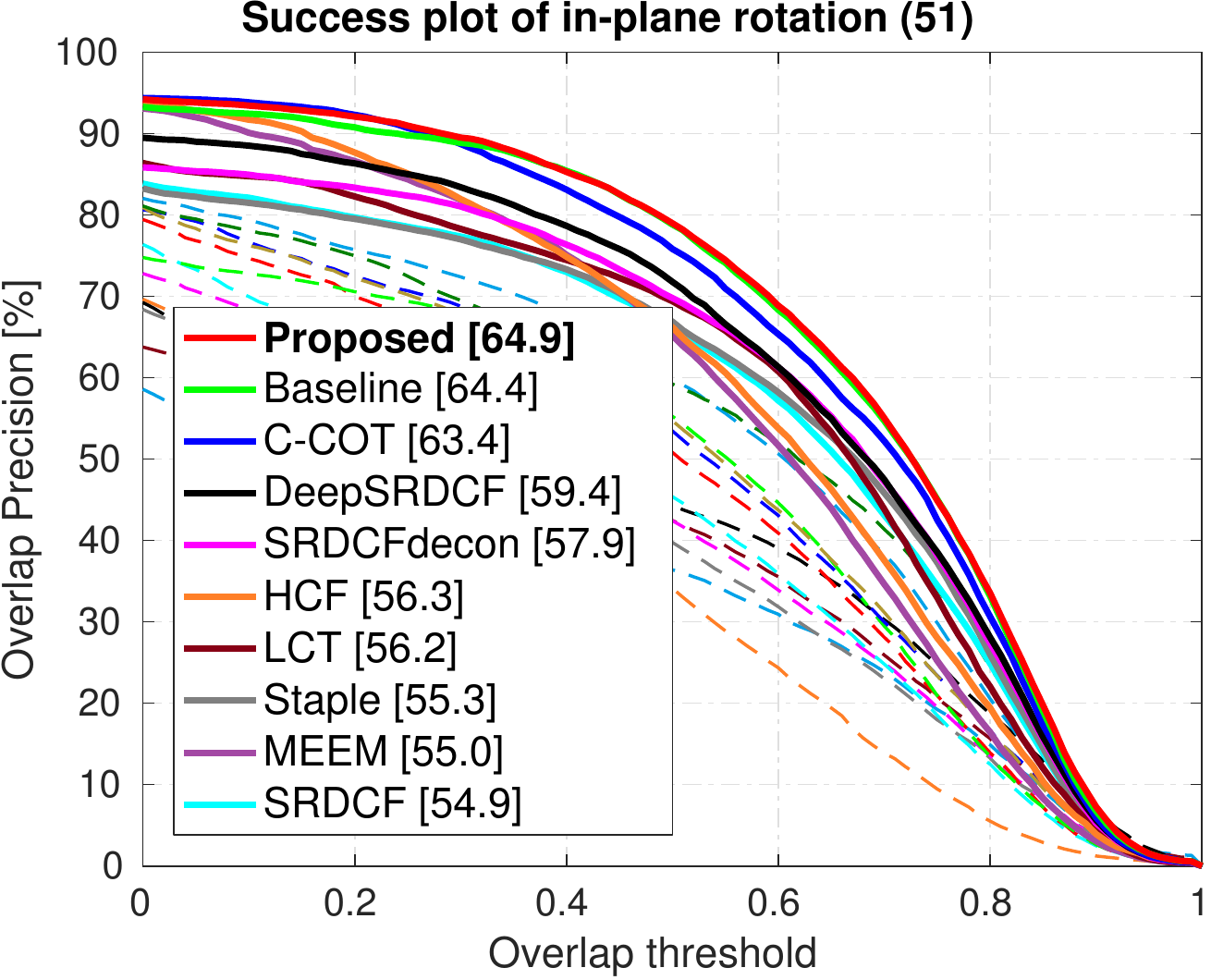}%
  \includegraphics[width=0.33\textwidth]{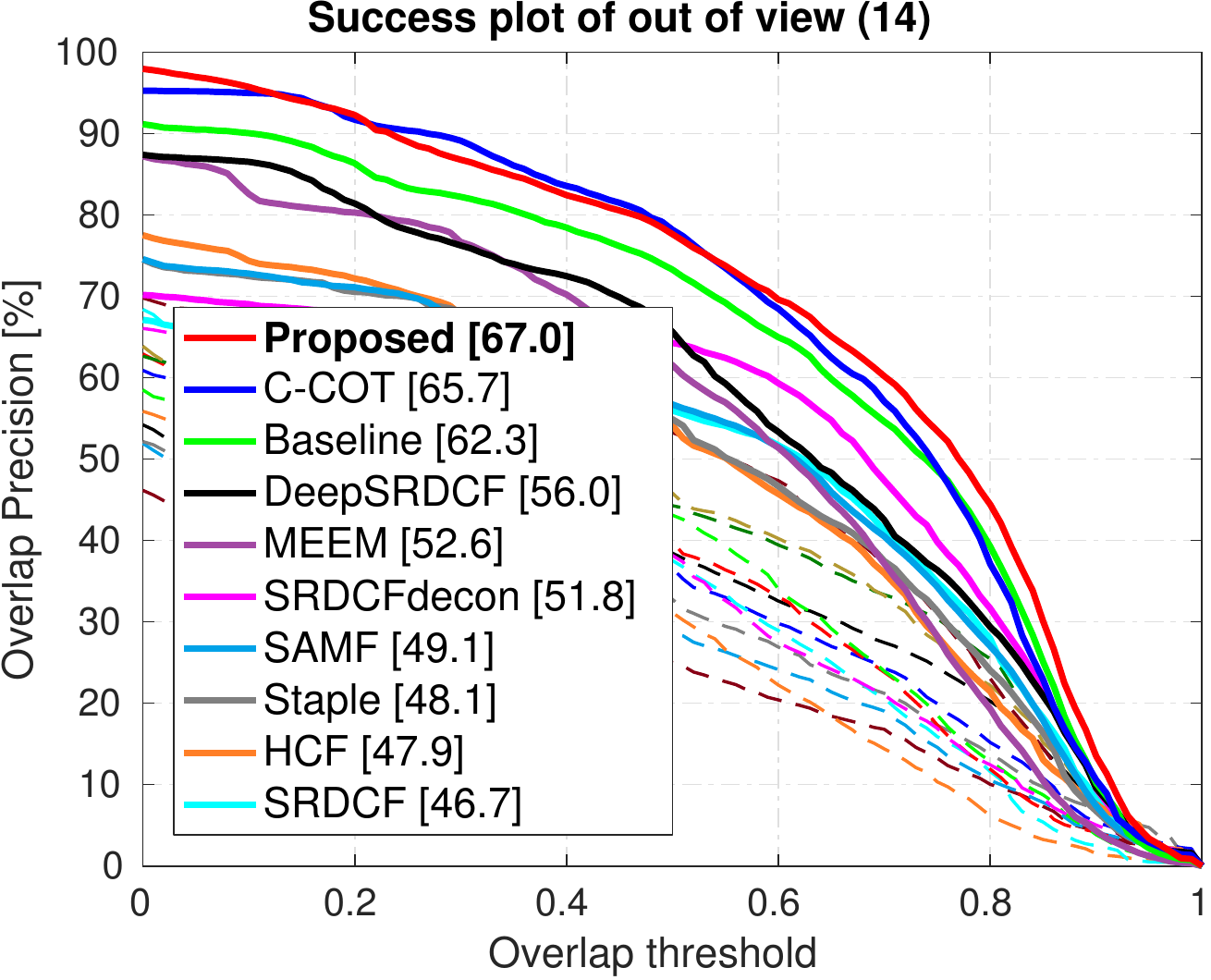}
  \includegraphics[width=0.33\textwidth]{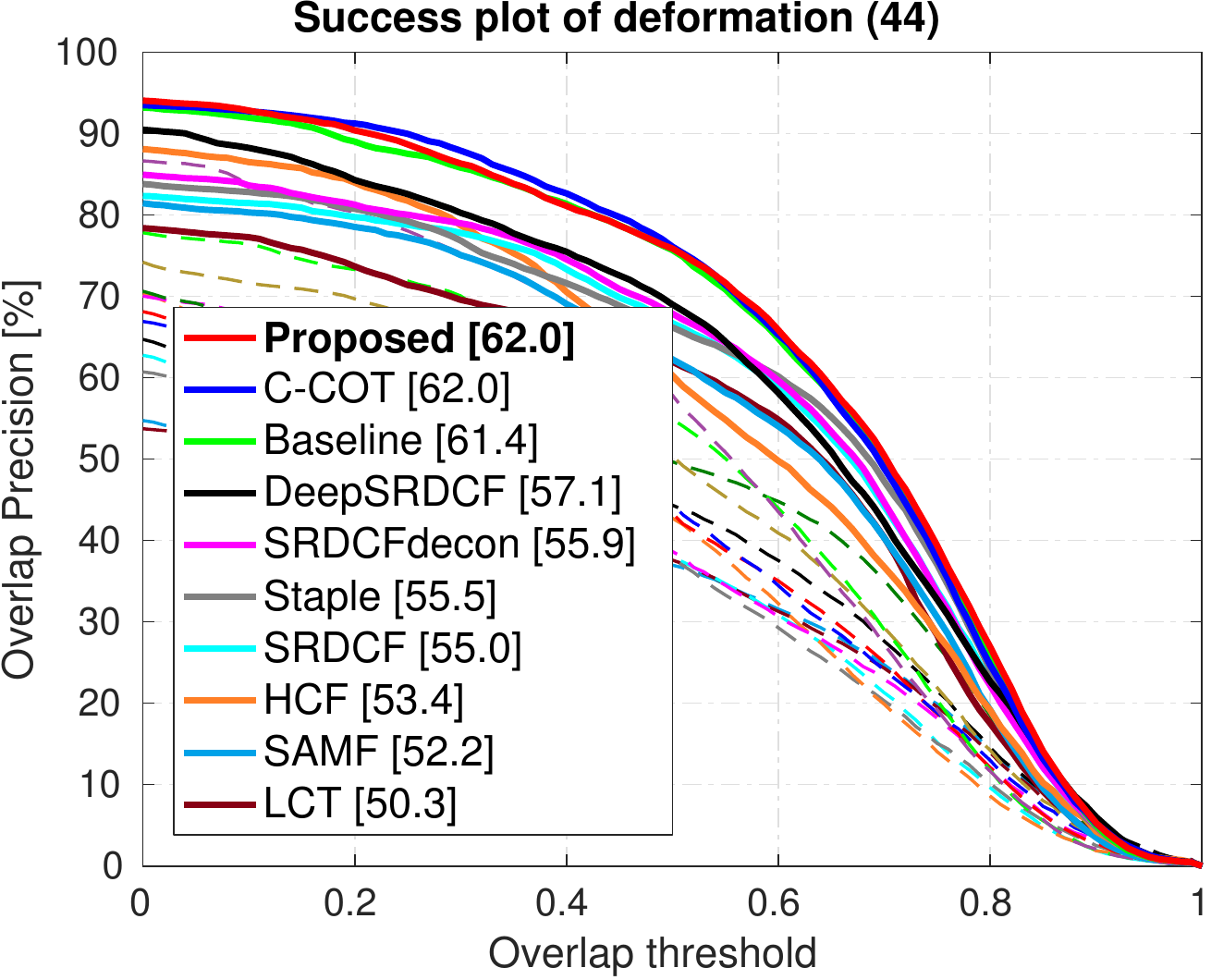}%
  \includegraphics[width=0.33\textwidth]{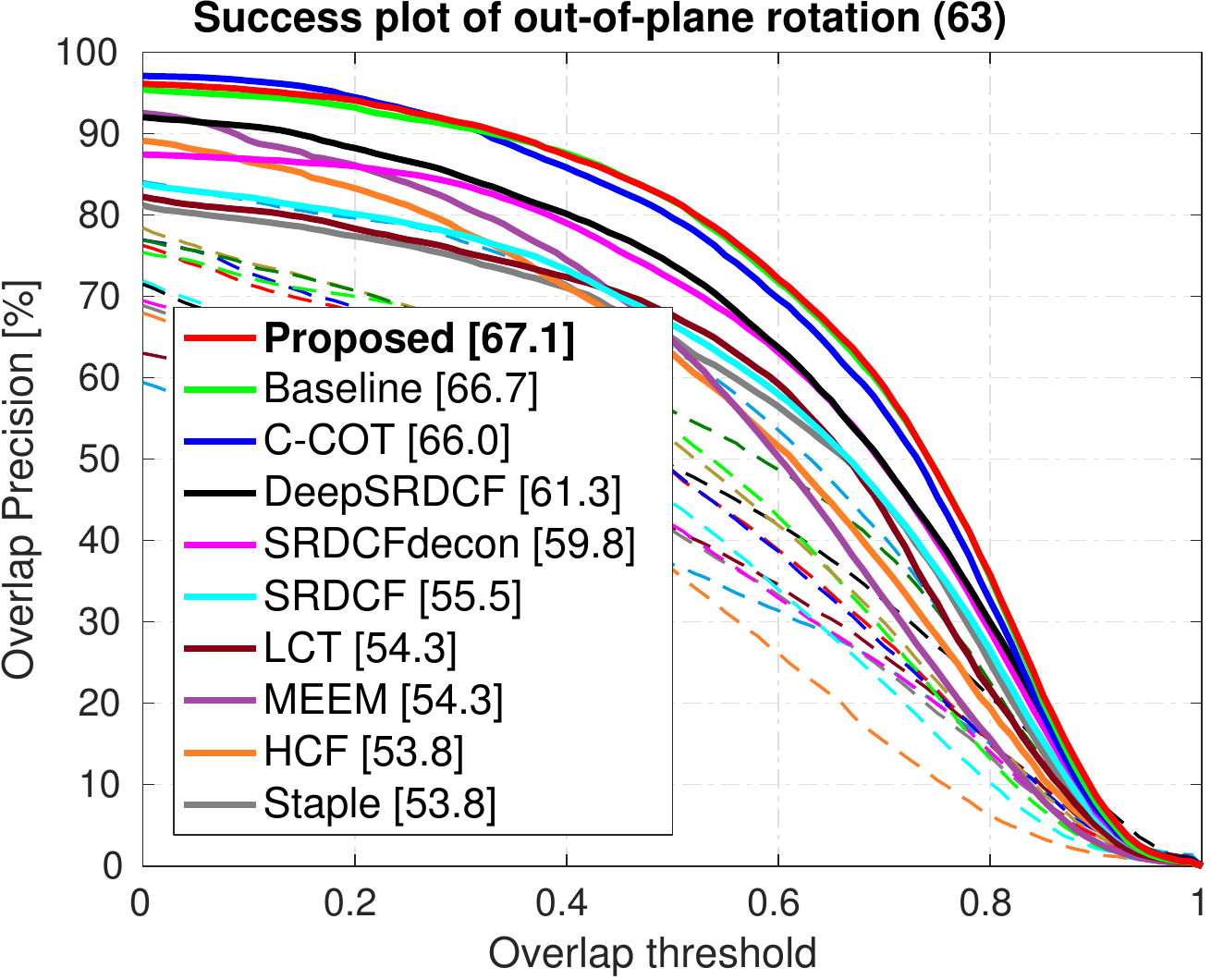}%
  \includegraphics[width=0.33\textwidth]{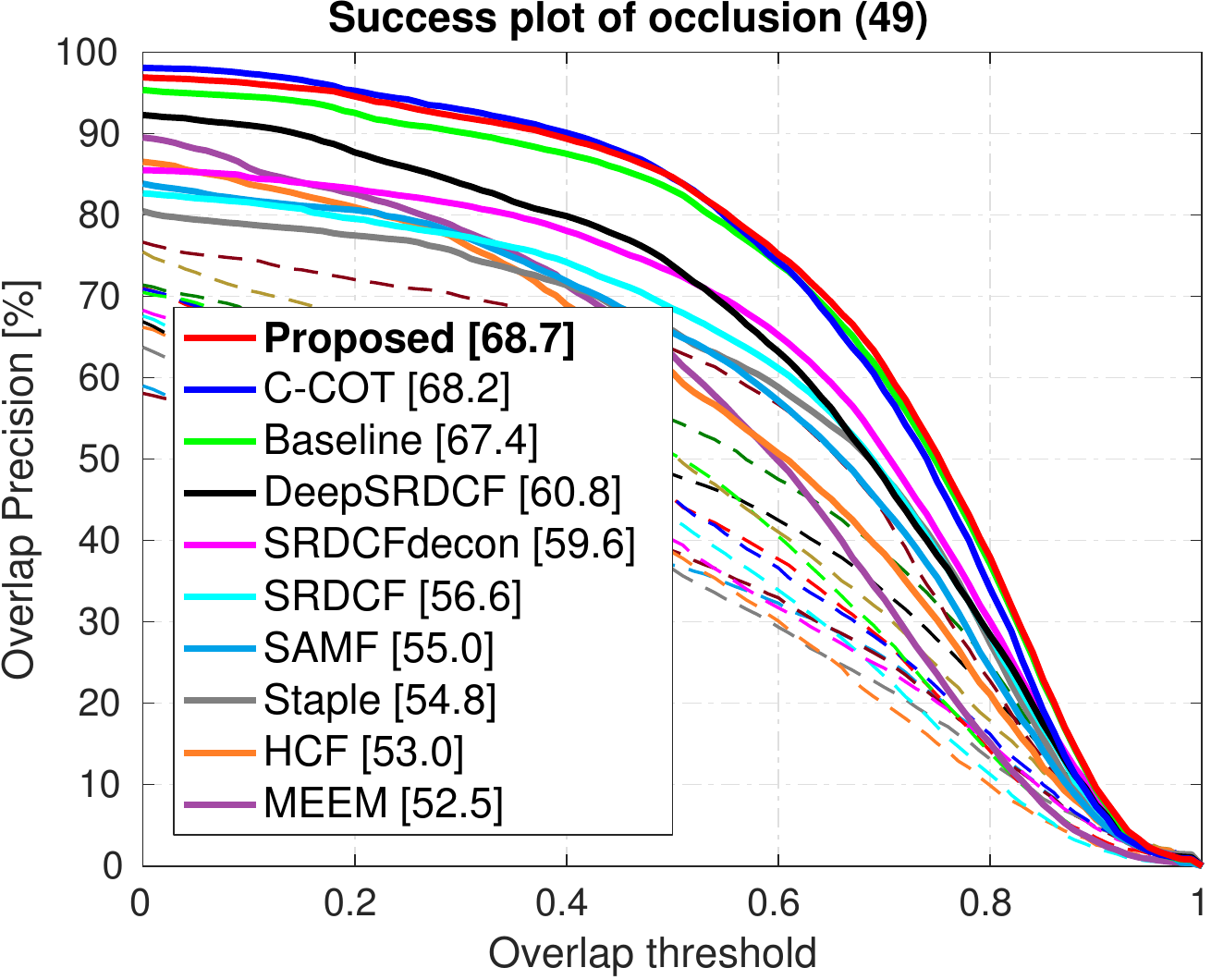}
  \caption{Attribute-based comparison on the OTB-2015 dataset. Success plots are shown for six attributes. Our approach achieves improved performance compared to existing trackers in these scenarios.}
  \label{fig:attribute}
\end{figure}

Table \ref{table:baselinetransf} shows the results of the baseline and proposed approach with the sub-filter positions regularized either with an affine transform, or the identity transform (Sec. 4.4). The proposed approach based on an affine transform provides improved tracking performance. This shows that regularization of the sub-filter positions is important and using an affine transform is superior compared to an identity transform. Table \ref{table:baselinefeat} shows the baseline comparison when using different set of features. The deep features provide improved performance. However, performance comparable to deep features is also achieved by using colornames.

\subsection{State-of-the-art Comparison}
\subsubsection{OTB-2015}
Figure \ref{fig:results} (on the left) shows the success plot for the OTB-2015 dataset which consists of 100 videos. The area-under-the-curve (AUC) score for each tracker is represented in the legend. Among existing approaches, the C-COT tracker \cite{DanelljanECCV2016} achieves an AUC score of $68.2\%$. It is worth to mention that the recently introduced ECO tracker \cite{DanelljanCVPR2017} achieves the best results with an AUC score of $70.0\%$. However, the ECO tracker also employs HOG features together with colornames (CN) and deep features. Instead, our deformable convolution filter approach achieves competetive performance without using HOG features, with an AUC score of $69.0\%$. Figure \ref{fig:attribute} shows the attribute based comparison on the OTB-2015 dataset. All videos in the OTB-2015 dataset are annotated with 11 different attributes. Our approach provides the best results on 7 attributes.

\subsection{TempleColor}
Figure \ref{fig:results} (on the right) shows the success plot for the TempleColor dataset consisting of 128 videos. The SRDCF tracker \cite{DanelljanICCV2015} and its deep features variant (DeepSRDCF) \cite{DanelljanVOT2015} achieve AUC scores of $51.6\%$ and $54.3\%$ respectively. The C-COT tracker yields an AUC score of $58.1\%$. Our approach improves the performance by $1.4\%$ compared to the C-COT tracker.

\subsection{VOT2016}
The VOT2016 which consists of 60 videos compiled from a set of more than 300 videos. On the VOT2016 dataset, the tracking performance is evaluated both in terms of accuracy (average overlap during successful tracking) and robustness (failure rate). The overall tracking performance is calculated using Expected Average Overlap (EAO) which takes into account both accuracy and robustness. For more details, we refer to \cite{VOT2015}. Table \ref{table:vot} shows the comparison on the VOT2016 dataset. We present the results in terms of EAO, failure rate, and accuracy. Our approach provides competetive performance in terms of accuracy and provides the best results in terms of robustness, with a failure rate of $0.70$.

\begin{table}[!t]
  \caption{State-of-the-art in terms of expected area overlap (EAO), robustness (failure rate), and accuracy on the VOT2016 dataset. The proposed approach show a slight decrease in EAO but a slight improvement to failure rate.}
  \label{table:vot}
  \resizebox{\columnwidth}{!}{%
    \begin{tabular}{l@{~}c@{~~}c@{~~}c@{~~}c@{~~}c@{~~}c@{~~}c@{~~}c@{~~}c@{~~}c@{~~}c}
      \toprule
      &SRBT&EBT&DDC&Staple&MLDF&SSAT&TCNN&C-COT&ECO&\textbf{Proposed}\\
      &\cite{VOT2016}&\cite{zhu2015tracking}&\cite{VOT2016}&\cite{Staple}&\cite{VOT2016}&\cite{VOT2016}&\cite{TCNN}&\cite{DanelljanECCV2016}&\cite{DanelljanCVPR2017}&Our\\\midrule
      EAO&0.290&0.291&0.293&0.295&0.311&0.321&0.325&0.331&\tabfirst{0.374}&\tabsecond{0.368}\\
      Fail.\ rt.&1.25&0.90&1.23&1.35&0.83&1.04&0.96&0.85&\tabsecond{0.72}&\tabfirst{0.70}\\
      Acc.&0.50&0.44&0.53&\tabsecond{0.54}&0.48&\tabfirst{0.57}&\tabsecond{0.54}&0.52&\tabsecond{0.54}&\tabsecond{0.54}\\\bottomrule
    \end{tabular}%
  }
  \vspace{-0.35cm}
\end{table}

\section{Conclusions}
We proposed a unified formulation to learn a deformable convolution filter. We represented our deformable filter as a linear combination of sub-filters. Both the coefficients and locations of all sub-filters are learned jointly in our framework. Experiments are performed on three challenging tracking datasets: OTB-2015, TempleColor and VOT2016. Our results clearly suggest that the proposed deformable convolution filter provides improved results compared to the baseline, leading to competitive performance compared to state-of-the-art trackers.

\noindent\textbf{Acknowledgments}:
This work has been supported by SSF (SymbiCloud), VR (EMC${}^2$, starting grant 2016-05543), SNIC, WASP, and Nvidia.

\bibliography{references}

\end{document}